\documentclass{article}

\usepackage{microtype}
\usepackage{graphicx}
\usepackage{subcaption}
\usepackage{booktabs} 
\usepackage[table]{xcolor} 
\usepackage{hyperref}
\usepackage{makecell}

\usepackage[accepted]{icml2026}

\usepackage{xspace}
\usepackage{amssymb} 
\usepackage{multirow}
\usepackage{array}
\usepackage[most]{tcolorbox}
\usepackage{listings}
\usepackage{graphicx} 
\newcolumntype{P}[1]{>{\centering\arraybackslash}p{#1}}

\lstdefinelanguage{json}{
    basicstyle=\ttfamily\small, 
    numbers=none, 
    showstringspaces=false, 
    breaklines=true, 
    morestring=[b]",
    morecomment=[l]{//},
    morecomment=[s]{/*}{*/},
    morekeywords={true,false,null}
}

\NewDocumentCommand{\heng}
{ mO{} }{\textcolor{red}{\textsuperscript{\textit{Heng}}\textsf{\textbf{\small[#1]}}}}
\definecolor{morandiGreen}{RGB}{80, 130, 100}
\definecolor{morandiYellow}{RGB}{200, 140, 80}
\definecolor{morandiBlue}{RGB}{70, 100, 140}
\definecolor{morandiPink}{RGB}{180, 80, 100}
\usepackage{amsmath}
\usepackage{amssymb}
\usepackage{mathtools}
\usepackage{amsthm}

\usepackage[capitalize,noabbrev]{cleveref}

\theoremstyle{plain}

\theoremstyle{definition}

\theoremstyle{remark}

\definecolor{bgcolor}{RGB}{56, 90, 70}

\newcommand{\name}[0]{ERGeoBench\xspace}
\usepackage[textsize=tiny]{todonotes}

\icmltitlerunning{A Comprehensive Benchmark for Embodied Reasoning and Geo-localization in Multimodal Large Language Models}

\begin{document}

\twocolumn[
  \icmltitle{\name: A Comprehensive Benchmark for Embodied Reasoning and Geo-localization in Multimodal Large Language Models}



  \icmlsetsymbol{equal}{*}
  \icmlsetsymbol{corr}{\dag}
  \begin{icmlauthorlist}
    \icmlauthor{Kaiwen Xue}{bupt}
    \icmlauthor{Tao Wei}{bupt}
    \icmlauthor{Guoxin Zhang}{bupt}
    \icmlauthor{Zhonghong Ou}{skl,corr}
    \icmlauthor{Kaoyan Lu}{sjtu}
    \icmlauthor{Yu Feng}{cmri}
    \icmlauthor{Yifan Zhu}{bupt}
    \icmlauthor{Haoran Luo}{ntu,corr}
  \end{icmlauthorlist}

  \icmlaffiliation{bupt}{Beijing University of Posts and Telecommunications, Beijing, China}
  \icmlaffiliation{skl}{State Key Laboratory of Networking and Switching Technology, Beijing University of Posts and Telecommunications, Beijing, China}
  \icmlaffiliation{sjtu}{School of Materials Science and Engineering, Shanghai Jiao Tong University, Shanghai, China}
  \icmlaffiliation{cmri}{China Mobile Research Institute, Beijing, China}
  \icmlaffiliation{ntu}{College of Computing and Data Science, Nanyang Technological University, Singapore}

\icmlcorrespondingauthor{Zhonghong Ou}{zhonghong.ou@bupt.edu.cn}
\icmlcorrespondingauthor{Haoran Luo}{haoran.luo@ieee.org}

  \icmlkeywords{Machine Learning, ICML}

  \vskip 0.3in
]



\printAffiliationsAndNotice{}  

\begin{abstract}
Multimodal large language models (MLLMs) have shown strong potential as embodied agents, yet embodied geo-localization remains underexplored due to the lack of fine-grained evaluation. We introduce \textbf{\name}, a diagnostic benchmark for vision-driven embodied geo-localization. \textbf{\name} evaluates models under three progressive settings---single-view, panorama-view, and embodied-view---where agents may actively acquire observations through sequential changes in yaw, pitch, and zoom. The benchmark contains 2,207 globally distributed street-view panoramas and measures four complementary capabilities: foundational perception, spatial awareness, common sense reasoning, and geo-localization reasoning. Evaluations of leading proprietary and open-source MLLMs show that current models can infer high-level geographic semantics, but still struggle with fine-grained perceptual operations, metric localization, and spatial consistency across views. We further observe that geo-localization is strongly correlated with the other capability dimensions, suggesting that accurate localization depends on integrated perception, spatial reasoning, and commonsense inference rather than isolated visual recognition. Overall, \textbf{\name} provides a unified framework for diagnosing and advancing human-like embodied geo-localization. Project Page: \url{https://kaixuewen.github.io/ERGeoBench/}
\end{abstract}

\begin{figure*}[th!]
\begin{center}
\includegraphics[width=0.98\linewidth]{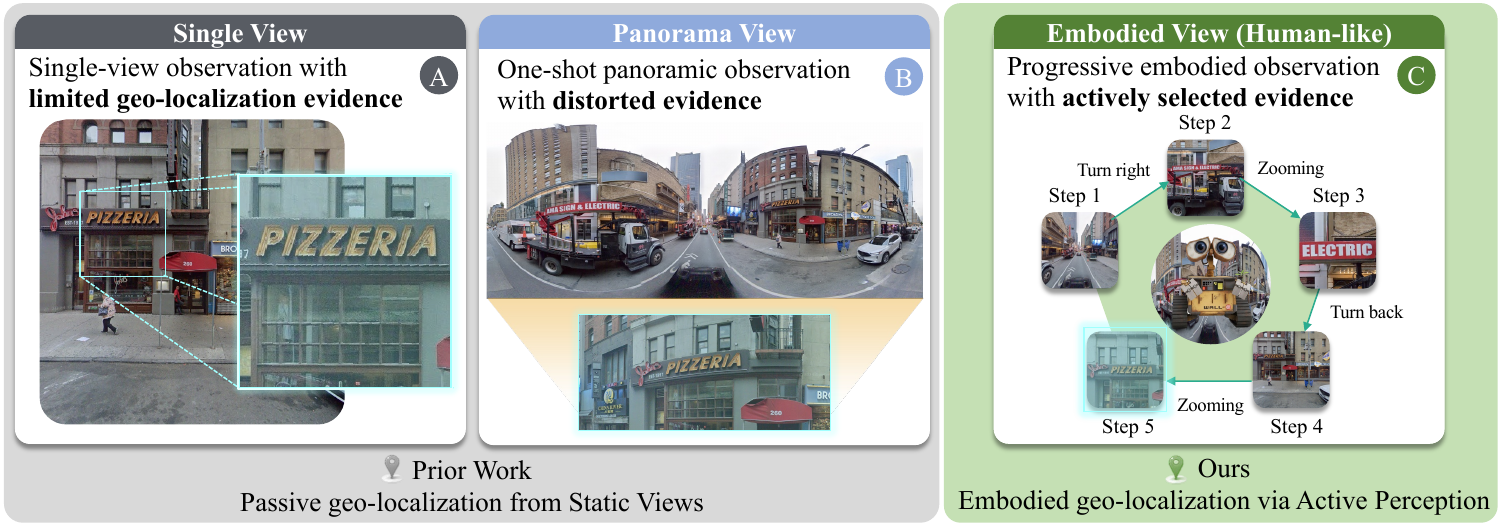}
\end{center}
\caption{Comparison of geo-localization paradigms under different visual settings. Existing approaches typically treat geo-localization as passive inference from either a single static image or a one-shot panoramic observation. Such settings cannot request additional evidence when visual cues are ambiguous. In contrast, we model the MLLM as an embodied agent that sequentially controls rotation, pitch, and zoom, actively acquiring complementary views to refine geo-localization predictions through embodied perception.
}
\label{fig:teaser}
\end{figure*}
\section{Introduction}

Worldwide image geo-localization \cite{gre2025,jia2024g3,pramanick2022world,vo2017revisiting} aims to infer the geographic location of an image captured anywhere on Earth. Unlike region-specific localization, which is constrained to predefined environments \cite{noh2017large,lee2022correlation,shao2023global}, global geo-localization requires models to reason over diverse visual cues, cultural context, architectural styles, and geographic priors at worldwide scale. Recent progress in multimodal large language models (MLLMs) has substantially improved image understanding  \cite{fu2025multimodal}, generation \cite{li2025exploring}, and cross-modal reasoning \cite{haocan,xue2025crebench}, motivating their use for real-world geo-localization \cite{gre2025,li2025recognition}.

Despite these advances, most existing geo-localization methods still formulate the task as a static recognition problem, where geographic attributes are inferred directly from a single image \cite{campos2025gaea} (Fig. \ref{fig:teaser}A) or a pre-captured panoramic observation \cite{zhang2025panoramic} (Fig. \ref{fig:teaser}B) via regression or retrieval-based mappings. Such formulations assume that sufficient localization cues are available in fixed visual inputs, leaving no mechanism for interaction or evidence refinement, and thus struggle to resolve ambiguity in visually complex scenes.

This static paradigm contrasts with human localization behavior. In the absence of maps, GPS, or external positioning systems \cite{mirowski2018learning}, humans localize by progressively observing the surroundings, selecting informative viewpoints, integrating evidence over time, and reasoning about spatial layout and commonsense context (Fig.~\ref{fig:teaser}C) \cite{tolman1948cognitive,epstein2017cognitive}. Localization is therefore naturally sequential: turning, zooming, and revisiting landmarks can reduce uncertainty when initial evidence is insufficient \cite{yangembodiedbench}. This view is consistent with active-vision and active-sampling accounts in cognitive science, where gaze and attention are guided by task goals, information gain, and curiosity-driven sampling rather than by passive stimulus exposure alone \cite{hayhoe2005eye,gottlieb2018towards}. It is also consistent with spatial-memory evidence that humans maintain scene, location, and heading representations across viewpoint changes \cite{moser2008place,taube2007head,steel2021network}. This discrepancy raises a central question: \textbf{can MLLMs perform global geo-localization through embodied and interactive visual reasoning that better reflects human localization behavior?}

To address this gap, we introduce \textbf{\name} (\textbf{E}mbodied \textbf{R}easoning \textbf{Geo}-localization \textbf{Bench}mark), the first benchmark that brings real-world panoramas into embodied simulation. \name leverages 2,207 globally distributed panoramas and transforms them into an egocentric perception setting via a controllable camera model, enabling agents to actively acquire views through yaw, pitch, and zoom actions without access to global maps or privileged viewpoints. Under this setting, geo-localization becomes a progressive reasoning process in which agents select informative viewpoints, accumulate visual evidence, and iteratively infer their location in a human-like manner. Importantly, the embodied-view setting is not a higher-cost duplicate of panorama-view evaluation: panorama-view measures passive reasoning with near-complete visual access, whereas embodied-view measures active evidence selection, cross-view memory, and action-conditioned hypothesis revision under partial observability.

\name adopts a task- and capability-oriented evaluation design. It evaluates models under three visual information conditions---single-view, panorama-view, and embodied-view geo-localization---covering increasingly rich perception and action settings. In addition, \name provides fine-grained evaluation across four capability dimensions: foundational perception, spatial awareness, commonsense reasoning, and geo-localization reasoning. This design supports diagnosis beyond coarse accuracy metrics and reveals strong correlations between geo-localization performance and the other three capabilities, suggesting that accurate localization emerges from integrated perception and reasoning rather than isolated pattern matching.

Using a unified embodied-agent protocol that combines egocentric observations, action history, and in-context examples, we evaluate nine leading proprietary and open-source MLLMs on \name. Our experiments reveal three consistent findings: (1) MLLMs perform well in high-level semantic geo-localization but remain weak at fine-grained metric localization; (2) maintaining spatial consistency across actively selected views is a major bottleneck; and (3) effective embodied reasoning benefits from accumulating visual evidence, highlighting the importance of realistic egocentric perception.

In summary, our contributions are threefold:
\begin{itemize}
\item We propose \name, the first benchmark to systematically evaluate embodied geo-localization reasoning from real-world panoramas, with explicit emphasis on active evidence acquisition, spatial-memory consistency, and hypothesis refinement across sequential egocentric views.
\item We introduce a fine-grained, capability-oriented evaluation framework spanning three task settings and four reasoning dimensions.
\item We conduct large-scale evaluations and analyses of state-of-the-art MLLMs, providing insights toward building more human-like location-aware embodied agents.
\end{itemize}

\begin{figure*}[th!]
\begin{center}
\includegraphics[width=0.98\linewidth]{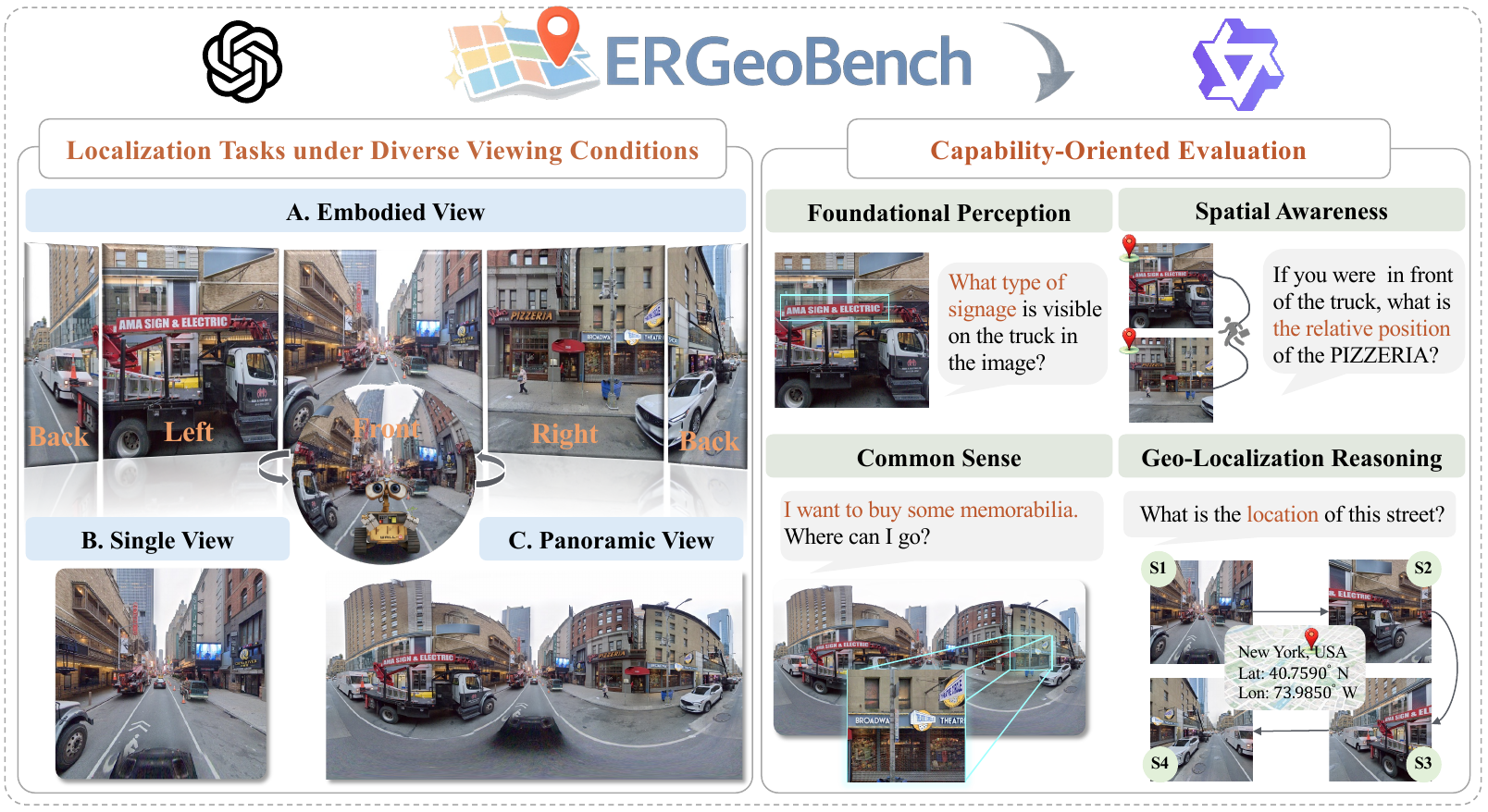}
\end{center}
\caption{Overview of ERGeoBench and its capability-oriented evaluation paradigm. ERGeoBench evaluates geo-localization under three visual information conditions: single-view, panorama-view, and embodied-view. Left: representative localization settings. Right: diagnostic evaluation over four core abilities---foundational perception, spatial awareness, commonsense reasoning, and geo-localization reasoning---using targeted visual questions grounded in the same environments.
}
\label{fig:Ergeobench}
\end{figure*}
\section{Related Work}

\noindent \textbf{Visual Geo-localization.} Traditional methods primarily treat geo-localization as a retrieval or classification problem. Retrieval-based frameworks match queries against geotagged databases but scale poorly in sparse regions, while classification-based approaches are limited by coarse granularity. Recently, MLLMs \cite{jia2024g3, gre2025, li2025recognition} have introduced a reasoning-driven paradigm, leveraging world knowledge to infer locations from subtle visual cues (e.g., architectural styles). However, these methods remain tethered to static, passive observation, lacking the dynamic capabilities for real-world interaction.

\noindent \textbf{Evaluation Benchmarks.} Geo-localization evaluation has shifted from simple distance metrics to diagnostic frameworks. IMAGEO-Bench \cite{li2025pixels} and GAEA-Bench \cite{campos2025gaea} systematically evaluate geospatial bias and conversational reasoning, with GAEA integrating OSM metadata for environmental grounding. Others, such as GRE Suite \cite{gre2025} and GLOBE \cite{li2025recognition}, emphasize reasoning chain quality through reinforcement learning (e.g., GRPO). While existing benchmarks demonstrate the cue-extraction ability of MLLMs, they focus on passive single-view inference. As shown in Table~\ref{tab:comparison}, ERGeoBench advances across single-view, panorama-view, and embodied-view settings, enabling active perception and progressive geo-localization through controllable exploration and multi-step evidence accumulation.

\noindent \textbf{Embodied Geo-localization and Our Positioning.} A critical gap exists in evaluating \textit{embodied} geo-localization, where human-like agents must actively explore—e.g., rotating to find signage or zooming to resolve visual ambiguity—to achieve precise localization. To bridge this, we introduce \name, a unified benchmark for MLLMs as embodied agents. Unlike prior static benchmarks, \name covers \textit{single-view}, \textit{panorama-view}, and \textit{active embodied-view} settings. Crucially, we decouple evaluation into four diagnostic dimensions: Foundational Perception (FP), Spatial Awareness (SA), Common Sense (CS), and Geo-localization Reasoning (GR). As shown in Fig.~\ref{fig:Ergeobench}, \name provides a controllable and perceptually realistic platform for researching location-aware agents.

\begin{table*}[t]
\centering
\caption{Comparison between \name and existing geo-localization benchmarks. \textbf{FP}, \textbf{SA}, \textbf{CS}, and \textbf{GR} denote \textbf{Foundational Perception}, \textbf{Spatial Awareness}, \textbf{Common Sense}, and \textbf{Geo-localization Reasoning}, respectively. $\checkmark$ and $\times$ indicate whether the feature is supported or not.}
\label{tab:comparison}

\resizebox{\textwidth}{!}{
\begin{tabular}{lccccc}
\toprule
Method & View & Active Perception & Embodied & Metric & Diagnostics \\ 
\midrule
IMAGEO-Bench & Single & $\times$ & $\times$ & $\times$ & GR \\
GLOBE        & Single & $\times$ & $\times$ & $\times$ & GR \\
GRE Suite    & Single & $\times$ & $\times$ & $\times$ & GR \\
GAEA         & Single & $\times$ & $\times$ & $\times$ & GR \\
G3           & Single & $\times$ & $\times$ & $\times$ & GR \\ 
\midrule
\textbf{ERGeoBench (Ours)} & Single / Panorama / Embodied & $\checkmark$ & $\checkmark$ & $\checkmark$ & FP \& SA \& CS \& GR \\ 
\bottomrule
\end{tabular}
}
\end{table*}

\begin{figure*}[th!]
\begin{center}
\includegraphics[width=0.98\linewidth]{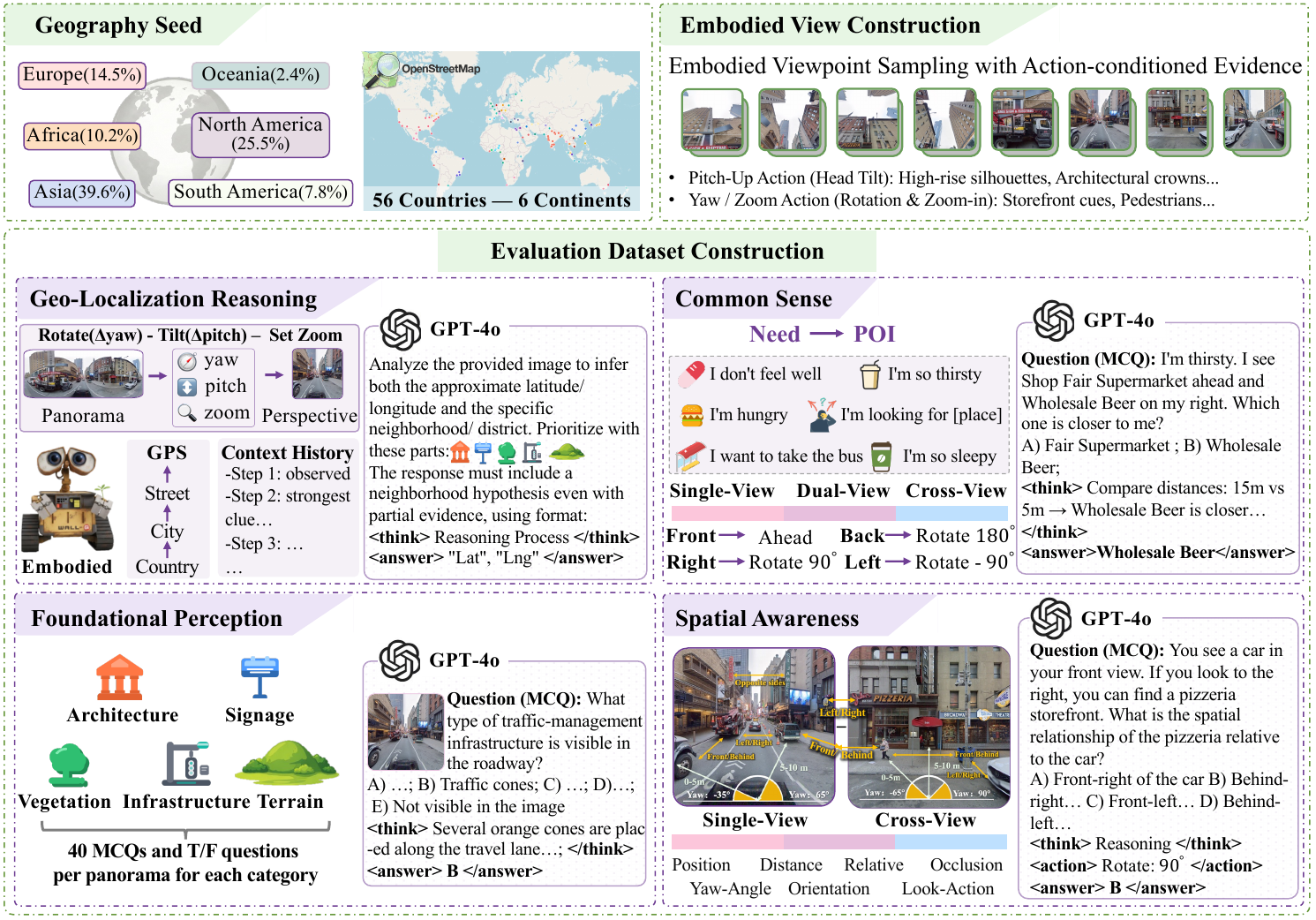}
\end{center}
\caption{ERGeoBench dataset construction pipeline. The pipeline integrates large-scale geographic data collection, embodied view construction via camera actions, and capability-oriented evaluation over geo-localization reasoning, common sense, foundational perception, spatial awareness.
}
\label{fig:dataset}
\end{figure*}

\section{Problem Formulation}
\label{sec:problem_formulation}

\subsection{Problem Setting}
We formulate the problem of \textbf{Embodied Reasoning Geo-localization} as a sequential decision-making process. Let $\mathcal{D} = \{(\mathcal{S}_i, \mathcal{Q}_i, \mathcal{A}_i)\}_{i=1}^N$ denote a dataset of $N$ scenarios sampled from diverse global urban environments. Each scenario $\mathcal{S}_i$ is represented by a high-resolution $360^\circ$ equirectangular panorama $P_i$, which serves as the latent environment state. Given a natural language query $\mathcal{Q}_i$, the model must generate a response $\hat{a}_i$ that matches the ground-truth answer $\mathcal{A}_i$ based on an observation sequence $\Omega_i$.

We evaluate model performance across four capability dimensions $\mathcal{C}$:
(i) \textbf{Foundational Perception} ($\mathcal{C}_{perc}$),
(ii) \textbf{Spatial Awareness} ($\mathcal{C}_{spat}$),
(iii) \textbf{Common Sense Reasoning} ($\mathcal{C}_{common}$), and
(iv) \textbf{Geo-localization Reasoning} ($\mathcal{C}_{geo}$).

\subsection{Visual Information Conditions}
\label{sec:visual_conditions}

We simulate a human-like embodied agent that interacts with the environment under an embodied multi-view setting. Unlike passive observation, the agent explores the panorama through a controllable action space defined by yaw ($\psi$), pitch ($\phi$), and zoom ($\zeta$):
\begin{equation}
    \mathcal{V} = \{v = (\psi, \phi, \zeta) \mid \psi \in \Psi, \phi \in \Phi, \zeta \in \mathcal{Z}\}
\end{equation}
where $\Psi$ represents the horizontal gaze direction, $\Phi$ represents the vertical inclination, and $\mathcal{Z}$ represents discrete zoom levels. The zoom level $\zeta$ inversely determines the field-of-view (FOV), allowing the agent to adaptively trade off between capturing global contextual cues with a wide FOV and inspecting fine-grained local details with a narrow FOV during active exploration.

For any action $v_t \in \mathcal{V}$ selected at step $t$, the visual observation $I_{v_t}$ is rendered via perspective projection from the latent panorama $P$:
\begin{equation}
\begin{split}
    I_{v_t} &= \text{Proj}(P; \psi_t, \phi_t, \text{FOV}(\zeta_t)), \\
    &\quad \text{with } \text{FOV}(\zeta) = \text{FOV}_{base} \cdot 2^{-\zeta}
\end{split}
\end{equation}
Here, $\zeta=1$ corresponds to the base FOV (e.g., $90^{\circ}$). The agent operates sequentially: at step $t$, given the interaction history $H_t = \{(v_\tau, I_{v_\tau})\}_{\tau=1}^t$, the policy $\pi(H_t, \mathcal{Q})$ decides whether to output a final geo-localization answer $\hat{a}_i$ or select the next viewpoint $v_{t+1}$ to acquire additional evidence. Thus, yaw, pitch, and zoom are treated as task-conditioned sensing actions rather than image augmentations: yaw changes the sampled sector, pitch targets overhead or street-level cues, and zoom resolves visible but unreadable details. The resulting observe--evaluate--update--act loop directly tests active sampling and cross-view memory.

\subsection{Evaluation Metrics}
\label{sec:metrics}
To evaluate the multifaceted capabilities of MLLMs in geo-localization, we adopt a hybrid evaluation strategy that measures both semantic classification accuracy for geographic attributes and metric localization precision for physical coordinates. To provide a unified ranking index, we introduce the \textbf{Geo-localization Score (GLS)}, a normalized metric ranging from 0 to 100.

Formally, GLS is defined as the arithmetic mean of three component scores, representing semantic understanding, hit-rate precision, and error magnitude, respectively:
\begin{equation}
    \text{GLS} = \frac{1}{3} \left( \mathcal{S}_{sem} + \mathcal{S}_{met} + \mathcal{S}_{err} \right)
\end{equation}

Each component is calculated as follows:

\noindent \textbf{Semantic Alignment Score ($\mathcal{S}_{sem}$).}
This component evaluates whether the model correctly predicts hierarchical geographic labels. Let $y_i^{(k)}$ be the ground-truth label at granularity level $k \in \{\text{street}, \text{city}, \text{country}\}$ and $\hat{y}_i^{(k)}$ be the predicted label parsed from the model response. The score is the average accuracy across these three levels:
\begin{equation}
    \mathcal{S}_{sem} = \frac{1}{3} \sum_{k} \left( \frac{1}{N} \sum_{i=1}^{N} \mathbb{I}(y_i^{(k)} = \hat{y}_i^{(k)}) \right) \times 100
\end{equation}
where $N$ is the total number of samples and $\mathbb{I}(\cdot)$ is the indicator function.

\noindent \textbf{Metric Precision Score ($\mathcal{S}_{met}$).}
This component evaluates physical proximity. First, we define the localization error $d(g_i, \hat{g}_i)$ between the ground-truth GPS $g_i=(\phi_i, \lambda_i)$ and the prediction $\hat{g}_i=(\hat{\phi}_i, \hat{\lambda}_i)$ using the Haversine formula:
\begin{equation}
\begin{split}
    d(g_i, \hat{g}_i) = 2R \cdot \arcsin \Bigg[ \bigg( \sin^2\left(\frac{\Delta\phi}{2}\right) & \\
    + \cos \phi_i \cos \hat{\phi}_i \sin^2\left(\frac{\Delta\lambda}{2}\right) \bigg)^{1/2} &\Bigg]
\end{split}
\end{equation}
where $\phi$ and $\lambda$ denote latitude and longitude in radians, $\Delta\phi = \hat{\phi}_i - \phi_i$, $\Delta\lambda = \hat{\lambda}_i - \lambda_i$, and $R \approx 6,371$ km is the Earth's mean radius.

Based on this geodesic distance, $\mathcal{S}_{met}$ is calculated as the average ``hit rate'' across five dynamic thresholds $\mathbb{T} = \{1, 25, 200, 750, 2500\} \text{ km}$. Following the multi-threshold design in nuScenes~\cite{caesar2020nuscenes}, this rewards models that achieve fine-grained localization while maintaining robustness at larger scales:
\begin{equation}
    \mathcal{S}_{met} = \frac{1}{|\mathbb{T}|} \sum_{\tau \in \mathbb{T}} \left( \frac{1}{N} \sum_{i=1}^{N} \mathbb{I}\left( d(g_i, \hat{g}_i) \le \tau \right) \right) \times 100
\end{equation}

\noindent \textbf{Error Penalization Score ($\mathcal{S}_{err}$).}
While $\mathcal{S}_{met}$ measures the success rate within thresholds, $\mathcal{S}_{err}$ evaluates the magnitude of deviations for the entire dataset. We utilize the \textbf{Median Distance Error (MdDE)}, denoted as $E_{med}$, instead of the Mean Error to ensure robustness against long-tail outliers (e.g., antipodal failures). 

To map the unbounded error to a normalized $[0, 100]$ scale, we apply logarithmic normalization. This reflects the heavy-tailed nature of geospatial errors, penalizing short-range deviations (e.g., 1km vs. 10km) more strictly than long-range fluctuations:
\begin{equation}
    \mathcal{S}_{err} = \max\left(0, \ 1 - \frac{\ln(E_{med} + 1)}{\ln(D_{max} + 1)}\right) \times 100
\end{equation}
where $D_{max}$ is set to half of Earth's circumference (20,037.5 km).

\section{ERGeoBench}
\label{sec:ergeobench}

To instantiate the problem formulation above, we introduce \name, a large-scale benchmark designed to evaluate MLLMs in realistic, globally diverse, and interactive environments, as shown in Fig.~\ref{fig:Ergeobench}.

\subsection{Task Settings Implementation}
We implement three information conditions using a unified panorama database to assess varying levels of agency. The \textbf{Single-view} setting serves as a static baseline, strictly limiting observation to a fixed $90^\circ$ FOV street-level perspective (Pitch $\approx 0^\circ$). The \textbf{Panorama-view} setting provides the full equirectangular projection, serving as a passive information upper bound with maximal evidence availability. In contrast, the \textbf{Embodied-view} setting enables active exploration via a continuous action space (Yaw, Pitch, Zoom) subject to realistic kinematic constraints: lateral rotations are restricted to $|\Delta\psi| \geq 45^\circ$ to prevent insignificant jitter, vertical observation is bounded to $\phi \in [-60^\circ, 60^\circ]$, and magnification is adjustable within $\zeta \in [1\times, 5\times]$, forcing agents to deliberatively trade off broad context for fine-grained details. This design makes panorama-view and embodied-view complementary: the former tests whether a model can exploit complete visual evidence, while the latter tests whether it can obtain and integrate the right evidence through sequential actions.

\subsection{Data Collection Pipeline}
Our data collection pipeline consists of three stages (as shown in Fig.~\ref{fig:dataset}): global sampling, diversity-aware filtering, and visual projection.

\paragraph{Global Sampling \& Filtering.}
We sourced candidate locations from the top 1,000 populated cities worldwide, excluding regions with restricted Street View coverage. To ensure high information density, we focused on a 5km radius around city centers and utilized OpenStreetMap (OSM) data. We applied a weighted scoring mechanism $P_r$ to select representative street nodes:
\begin{equation}
P_r = w_d \cdot S_d(r) + w_l \cdot S_l(r)
\end{equation}
where $S_d(r)$ prioritizes roads closer to the city center, and $S_l(r)$ favors major arterial roads over backstreets. We set $w_d=0.7$ and $w_l=0.3$ to balance centrality and road significance.

\paragraph{High-Fidelity Projection.}
To minimize edge distortion common in $360^\circ$ imagery, we established a precise mapping model between spherical coordinates and perspective FOV images. We rendered standard views with a fixed focal length to strictly simulate human visual observation, eliminating the ``fisheye'' artifacts often present in raw panoramic data.

\subsection{Capability-Oriented Evaluation}
\label{sec:capabilities}

\name evaluates models across four hierarchical capabilities, moving from local perception to global reasoning.

\paragraph{Geo-localization Reasoning.}
As the integrative capability, geo-localization reasoning requires the model to synthesize foundational perception, spatial awareness, and commonsense reasoning to infer geographic attributes such as country, city, street-level location, and GPS coordinates. This moves beyond static pattern matching toward evidence-based localization.

\paragraph{Foundational Perception.}
This dimension assesses fine-grained visual recognition via a 5-Slot Framework (Architecture, Signage, Vegetation, Terrain, Infrastructure). Task Design: We generate Multiple Choice and True/False questions targeting specific FOVs. Quality Control: We use Target View Declaration to bind each question to a visible region, then apply layered verification rather than relying on a single model as ground truth: candidate annotations are checked by an independent VLM and manually verified by the authors, while inconsistent or ambiguous samples are corrected, discarded, or regenerated. This design reduces hallucination and mitigates potential self-referential bias from model-assisted annotation.

\paragraph{Spatial Awareness.}
This dimension evaluates egocentric geometric reasoning across 7 sub-capabilities. We first assess the model's grasp of intrinsic geometric properties through \textit{Yaw Estimation}, \textit{2D Positioning}, \textit{Depth Estimation}, and \textit{Distance} perception. Beyond absolute metrics, we evaluate relative spatial understanding via \textit{Left/Right} and \textit{Front/Back} relational reasoning tasks, which require the agent to ground objects within its egocentric frame. A critical contribution is the \textit{Cross-view Reasoning} task, where the model must deduce an object's position relative to a target view after a rotation (e.g., ``If I turn 90 degrees right, where is the object?''). This explicitly tests the agent's ability to maintain a consistent 3D mental map across viewpoint changes.

\paragraph{Common-sense Reasoning.}
This dimension focuses on Functional Affordance Alignment. Models must identify Points of Interest (POIs) that satisfy abstract user needs (e.g., ``I am thirsty'' $\to$ Convenience Store). \textbf{Hierarchical Complexity:} Tasks range from \textit{Single-View} identification to \textit{Dual-View} combination (street-level entrances + skyline signage) and \textit{Cross-View} accessibility comparison. \textbf{Embodied Integration:} The tasks explicitly require models to infer visibility and reachability, simulating a pedestrian deciding where to go based on visual evidence. For reachability-style questions, ground truth is represented by discrete POI labels or constrained comparative answers, avoiding subjective continuous thresholds while preserving the functional reasoning signal.

\begin{table*}[t]
\centering
\caption{Performance comparison of proprietary and open-source MLLMs on our \name. The evaluation covers three distinct settings: Single-view, Panorama-view, and Embodied-view Geolocation. We report hierarchical classification accuracy (Street to Country), distance-based accuracy (1km to 2500km), and error metrics (Average and Median). ``$\uparrow$'' indicates higher is better, and ``$\downarrow$'' indicates lower is better. The best results are highlighted in \textbf{bold}.}
\label{tab:main_results}
\resizebox{1\linewidth}{!}{
\begin{tabular}{llccccccccc}
\toprule
\multicolumn{2}{c}{\multirow{2}{*}[-0.3em]{\textbf{Model}}} & \multicolumn{5}{c}{\textbf{Proprietary MLLMs}} & \multicolumn{4}{c}{\textbf{Open-Source MLLMs}} \\ 
\cmidrule(lr){3-7} \cmidrule(lr){8-11} 
\multicolumn{2}{c}{} & GPT-4o & \makecell{Gemini-2.0\\Flash} & \makecell{Gemini-2.5\\Pro} & \makecell{Gemini-3\\Flash} & \makecell{Qwen-VL\\Max} & \makecell{Qwen2.5\\-7B} & \makecell{Qwen3\\-8B} & \makecell{InternVL\\2.5-8B} & \makecell{InternVL\\3-8B} \\
\midrule

\multirow{10}{*}{\rotatebox{90}{Single Geo}} 
& Street (\%) $\uparrow$ & 1.86 & 3.03 & 4.79 & \textbf{5.36} & 0.32 & 1.93 & 0.82 & 1.41 & 1.00 \\
& City (\%) $\uparrow$ & 46.21 & 48.65 & 48.92 & \textbf{50.54} & 21.71 & 17.61 & 16.98 & 8.93 & 25.42 \\
& Country (\%) $\uparrow$ & 84.16 & 87.37 & \textbf{94.28} & 80.09 & 66.35 & 45.40 & 71.07 & 48.72 & 68.82 \\
& Acc.@1km (\%) $\uparrow$ & 4.99 & 6.46 & 11.50 & \textbf{15.01} & 5.25 & 1.61 & 7.36 & 3.74 & 4.69 \\
& Acc.@25km (\%) $\uparrow$ & 63.91 & 69.02 & 70.74 & \textbf{73.99} & 37.95 & 27.73 & 36.38 & 19.85 & 42.07 \\
& Acc.@200km (\%) $\uparrow$ & 68.85 & 76.16 & 80.53 & \textbf{81.70} & 42.20 & 30.73 & 41.37 & 21.39 & 45.53 \\
& Acc.@750km (\%) $\uparrow$ & 85.43 & 91.29 & 92.76 & \textbf{93.83} & 62.69 & 39.56 & 64.12 & 35.04 & 59.17 \\
& Acc.@2500km (\%) $\uparrow$ & 94.38 & 97.80 & 98.24 & \textbf{98.66} & 82.23 & 50.48 & 84.56 & 53.08 & 71.39 \\
& Avg. Err (km) $\downarrow$ & 662.18 & 309.94 & 260.11 & \textbf{202.41} & 1805.14 & 2661.96 & 1600.29 & 3959.31 & 2150.07\\
& Med. Err (km) $\downarrow$ & 7.53 & \textbf{7.00} & 8.09 & 7.29 & 402.19 & 522.92 & 403.42 & 1237.39 & 330.28 \\ 
\rowcolor{gray!12} \cellcolor{white}
& \textbf{GLS} $\uparrow$ & 61.98 & 64.50 & \textbf{65.93} & 65.54 & 38.32 & 29.49 & 38.59 & 24.80 & 39.24 \\
\midrule

\multirow{10}{*}{\rotatebox{90}{Panorama}} 
& Street (\%) $\uparrow$ & 2.59 & 5.04 & 5.29 & \textbf{6.45} & 0.63 & 1.05 & 1.00 & 1.59 & 1.63 \\
& City (\%) $\uparrow$ & 54.03 & 50.93 & 52.03 & \textbf{54.51} & 29.53 & 26.69 & 28.20 & 14.11 & 28.52 \\
& Country (\%) $\uparrow$ & 92.71 & 89.23 & \textbf{96.33} & 81.83 & 75.44 & 58.14 & 84.88 & 68.06 & 75.55 \\
& Acc.@1km (\%) $\uparrow$ & 7.58 & 7.15 & 15.97 & \textbf{20.87} & 7.28 & 2.90 & 9.51 & 5.22 & 13.64 \\
& Acc.@25km (\%) $\uparrow$ & 75.50 & 72.82 & 75.45 & \textbf{78.21} & 47.44 & 38.11 & 51.65 & 27.72 & 47.21 \\
& Acc.@200km (\%) $\uparrow$ & 80.68 & 80.26 & 84.91 & \textbf{85.98} & 52.46 & 41.11 & 57.63 & 30.85 & 51.46 \\
& Acc.@750km (\%) $\uparrow$ & 93.11 & 93.54 & 96.03 & \textbf{96.12} & 72.46 & 48.97 & 79.76 & 49.50 & 65.01 \\
& Acc.@2500km (\%) $\uparrow$ & 98.78 & 98.82 & 99.51 & \textbf{99.80} & 88.56 & 58.99 & 94.39 & 68.42 & 75.29\\
& Avg. Err (km) $\downarrow$ & 231.51 & 205.67 & 140.00 & \textbf{107.50} & 1164.69 & 1738.76 & 700.96 & 3088.62 & 1796.13 \\
& Med. Err (km) $\downarrow$ & 6.06 & 6.52 & 6.81 & \textbf{5.71} & 121.74 & 18.35 & 15.09 & 760.40 & 176.95 \\ 
\rowcolor{gray!12} \cellcolor{white}
& \textbf{GLS} $\uparrow$ & 67.06 & 66.18 & \textbf{68.28} & 68.19 & 46.76 & 45.58 & 56.19 & 32.43 & 44.48 \\
\midrule

\multirow{10}{*}{\rotatebox{90}{Embodied}} 
& Street (\%) $\uparrow$ & 4.30 & 5.77 & 6.40 & \textbf{8.83} & 0.50 & 2.24 & 1.10 & 1.69 & 2.66 \\
& City (\%) $\uparrow$ & \textbf{55.08} & 52.69 & 53.45 & 53.41 & 24.98 & 19.85 & 22.45 & 16.63 & 29.76 \\
& Country (\%) $\uparrow$ & 92.33 & 90.22 & \textbf{99.01} & 81.24 & 73.44 & 48.66 & 78.10 & 66.56 & 74.22 \\
& Acc.@1km (\%) $\uparrow$ & 9.73 & 9.09 & 16.75 & \textbf{22.96} & 7.24 & 2.29 & 8.69 & 4.79 & 7.62 \\
& Acc.@25km (\%) $\uparrow$ & 76.10 & 74.34 & 74.38 & \textbf{79.14} & 43.67 & 30.95 & 44.03 & 28.56 & 46.68 \\
& Acc.@200km (\%) $\uparrow$ & 80.99 & 81.92 & 84.98 & \textbf{85.62} & 48.55 & 34.53 & 49.29 & 32.17 & 50.66 \\
& Acc.@750km (\%) $\uparrow$ & 93.11 & 95.01 & \textbf{98.03} & 95.63 & 67.96 & 43.58 & 71.70 & 51.20 & 65.10 \\
& Acc.@2500km (\%) $\uparrow$ & 98.63 & 99.12 & \textbf{99.26} & 98.82 & 86.92 & 54.13 & 89.44 & 72.05 & 76.71 \\
& Avg. Err (km) $\downarrow$ & 224.05 & 174.14 & \textbf{110.30} & 152.11 & 1336.23 & 2256.78 & 1147.53 & 2675.72 & 1787.16 \\
& Med. Err (km) $\downarrow$ & 5.99 & 6.30 & 6.67 & \textbf{4.85} & 261.18 & 269.64 & 229.04 & 666.55 & 192.68 \\ 
\rowcolor{gray!12} \cellcolor{white}
& \textbf{GLS} $\uparrow$ & 67.55 & 67.13 & \textbf{69.02} & 68.81 & 42.54 & 33.38 & 43.87 & 33.46 & 43.91 \\
\bottomrule
\end{tabular}}
\end{table*}

\begin{table*}[!t]
\centering
\caption{Fine-grained capability assessment across three dimensions: Foundational Perception, Spatial Awareness, and Common Sense. We report the accuracy (\%) for each sub-category. Abbreviations: Yaw Est. (Yaw Estimation), 2D Pos. (2D Positioning), L/R (Left/Right Relative), and F/B (Front/Back Relative). The best performance in each row is highlighted in \textbf{bold}.}
\label{tab:fine_grained}
\resizebox{1\linewidth}{!}{
\begin{tabular}{llccccccccc}
\toprule
\multicolumn{2}{c}{\multirow{2}{*}[-1em]{\textbf{Dimension}}} & \multicolumn{5}{c}{\textbf{Proprietary MLLMs}} & \multicolumn{4}{c}{\textbf{Open-Source MLLMs}} \\ 
\cmidrule(lr){3-7} \cmidrule(lr){8-11}
\multicolumn{2}{c}{} & GPT-4o & \makecell{Gemini-3\\Flash} & \makecell{Gemini-2.0\\Flash} & \makecell{Gemini-2.5\\Pro} & \makecell{Qwen-VL\\Max} & \makecell{Qwen2.5\\-VL-7B} & \makecell{Qwen3\\-VL-8B} & \makecell{InternVL\\2.5-8B} & \makecell{InternVL\\3-8B} \\ 
\midrule

\multirow{6}{*}{\textbf{\makecell[l]{Foundational\\Perception}}} 
& Architecture & \textbf{76.30} & 68.29 & 72.68 & 73.78 & 71.85 & 54.52 & 65.68 & 60.57 & 64.06 \\
& Infrastructure & \textbf{85.70} & 72.30 & 70.99 & 64.42 & 81.96 & 57.16 & 75.19 & 75.59 & 74.30 \\
& Signage & 77.93 & 61.23 & 60.34 & 54.44 & \textbf{78.17} & 59.84 & 63.71 & 68.74 & 70.63 \\
& Terrain & 93.49 & 59.11 & 67.33 & 83.44 & 93.84 & 73.94 & 79.52 & 75.74 & \textbf{94.03} \\
& Vegetation & \textbf{80.46} & 70.94 & 74.08 & 75.21 & 79.69 & 62.13 & 67.05 & 66.60 & 76.93 \\
\cmidrule(l){2-11} 
\rowcolor{gray!12} \cellcolor{white}
& \textit{Avg.} & \textbf{82.78} & 66.37 & 69.08 & 70.26 & 81.10 & 61.52 & 70.23 & 69.45 & 75.99 \\ 
\midrule

\multirow{8}{*}{\textbf{\makecell[l]{Spatial\\Awareness}}} 
& Yaw Est. & 31.24 & 12.84 & 39.91 & 14.83 & 33.76 & 24.15 & \textbf{40.32} & 28.42 & 33.81 \\
& 2D Pos. & 24.87 & 17.92 & \textbf{29.73} & 19.56 & 21.84 & 9.22 & 19.68 & 3.50 & 9.77 \\
& Depth Est. & 59.68 & 44.67 & 64.58 & 49.72 & 62.91 & 55.74 & \textbf{64.73} & 60.78 & 54.03 \\
& Distance & 55.43 & 59.38 & \textbf{69.82} & 56.28 & 63.58 & 59.02 & 64.89 & 55.21 & 58.47 \\
& L/R Relative & 64.92 & 59.72 & \textbf{79.36} & 69.45 & 77.69 & 72.63 & 77.54 & 67.31 & 66.71 \\
& F/B Relative & 53.76 & 54.81 & \textbf{69.54} & 59.81 & 66.83 & 62.50 & 66.71 & 60.10 & 63.82 \\
& Cross View & 17.15 & 9.87 & 14.89 & 11.67 & 15.92 & 17.08 & 13.85 & \textbf{20.83} & 18.69 \\
\cmidrule(l){2-11}
\rowcolor{gray!12} \cellcolor{white}
& \textit{Avg.} & 43.86 & 37.03 & \textbf{52.55} & 40.19 & 48.93 & 42.91 & 49.67 & 42.31 & 43.61 \\ 
\midrule

\multirow{4}{*}{\textbf{\makecell[l]{Common\\Sense}}} 
& Single-View & 55.46 & 51.78 & \textbf{96.31} & 52.87 & 33.61 & 59.70 & 79.92 & 51.23 & 65.98 \\
& Dual-View & 38.38 & 31.13 & 49.44 & 32.06 & 34.11 & 50.37 & \textbf{53.62} & 47.77 & 55.76 \\
& Cross-View & 33.31 & 28.15 & 36.40 & 29.78 & 31.29 & \textbf{41.57} & 32.98 & 39.66 & 36.80 \\
\cmidrule(l){2-11}
\rowcolor{gray!12} \cellcolor{white}
& \textit{Avg.} & 42.38 & 37.02 & \textbf{60.72} & 38.24 & 33.00 & 50.55 & 55.51 & 46.22 & 52.85 \\ 
\bottomrule
\end{tabular}}
\end{table*}

\section{Experiments}
\label{sec:experiments}
\subsection{Experimental Setups}

\textbf{Model Selection.} We benchmark 9 representative MLLMs to ensure a comprehensive evaluation, comprising 5 proprietary and 4 state-of-the-art (SOTA) open-source models. The proprietary models include GPT-4o \cite{OpenAI2024GPT4o}, the Gemini series (Gemini-2.0-Flash, 2.5-Pro, and 3-Flash) \cite{Google2024, Google2025, Google2025_G3}, and Qwen-VL-Max \cite{Bai2023}. The open-source group consists of the Qwen series (Qwen2.5-VL-7B, Qwen3-VL-8B) \cite{bai2025qwen2, Bai2025} and the InternVL series (InternVL2.5-8B, InternVL3-8B) \cite{Chen2025, Zhu2025}. All models are configured with a temperature of $0.1$ and a maximum generation length of $4,096$ tokens to ensure deterministic and complete responses. All settings use a unified structured prompt and JSON schema. The full prompt is provided in Appendix~\ref{app:prompt_design}.

\subsection{Benchmark Results}

\paragraph{Overall Results.}
As shown in Table~\ref{tab:main_results}, there is a clear performance gap between proprietary and open-source models. Proprietary models, particularly the Gemini series and GPT-4o, consistently achieve GLS scores above 60.0, whereas open-source models generally lag behind. This gap suggests that global geo-localization is knowledge-intensive and benefits from broad pretraining, robust visual perception, and instruction-following ability. Among all evaluated models, Gemini-2.5-Pro achieves the highest overall GLS across the three settings, while Gemini-3-Flash leads many fine-grained distance-threshold metrics, establishing strong baselines for embodied geo-localization agents.

\paragraph{Impact of Active Exploration.}
Comparing performance across the three task settings reveals the efficacy of embodied reasoning. The Panorama setting, which provides an omniscient field-of-view, consistently yields higher performance than the restricted Single-view baseline, serving as an information upper bound. Notably, we observe that capable embodied agents can bridge this information gap. Top-performing models like Gemini-3-Flash achieve GLS scores in the Embodied setting that approach or even surpass their Panorama performance. This suggests that through intelligent viewpoint selection, agents can effectively gather salient visual cues to compensate for restricted initial observations. The gain is most informative at fine-grained localization thresholds: for strong proprietary models, embodied exploration improves Acc.@1km over panorama-view evaluation (Appendix~\ref{app:embodied_vs_panorama}), indicating that sequential evidence accumulation can help resolve street-level ambiguity even when a panorama contains the same global scene. However, this active exploration capability is non-trivial; weaker models exhibit performance degradation in the Embodied setting compared to Panorama, likely due to disorientation caused by ineffective navigation actions.

\paragraph{Semantic Understanding vs. Metric Precision.}
A fine-grained analysis of the capability dimensions exposes a divergence between high-level semantic reasoning and low-level metric localization. While leading models achieve high accuracy in hierarchical classification (e.g., country-level inference), their performance drops sharply at fine-grained distance thresholds (e.g., 1km accuracy). This indicates that while MLLMs have mastered regional inference, precise coordinate regression remains a significant challenge. Furthermore, the substantial discrepancy between average and median errors across all models confirms the heavy-tailed nature of geospatial prediction errors, where antipodal failures are common. This validates the necessity of our GLS metric, which utilizes log-normalization to provide a robust assessment of model reliability against such outliers.


\subsection{Fine-grained Capability Analysis}
\label{sec:fine_grained_analysis}

To identify the specific strengths and weaknesses of current MLLMs, we perform a diagnostic evaluation across three foundational dimensions: Perception, Spatial Awareness, and Common Sense (as shown in Table~\ref{tab:fine_grained}).

\paragraph{Foundational Perception: A Solved Problem?}
The results in Table~\ref{tab:fine_grained} indicate that current MLLMs possess robust visual recognition capabilities. Most proprietary models achieve average scores above 70.0, with GPT-4o leading in complex categories such as Infrastructure (85.70) and Vegetation (80.46). Even open-source models like InternVL3 demonstrate competitive performance, particularly in Terrain classification (94.03). This suggests that the foundational ``what is in the image'' problem is largely solved for high-quality street views, serving as a reliable basis for downstream reasoning tasks. To check that this conclusion is not an artifact of GPT-4o-assisted annotation, Appendix~\ref{app:gpt_bias_check} reports additional GPT-5 and o1 evaluations; GPT-5 outperforms GPT-4o across all three capability dimensions, which weakens a simple self-preference explanation.

\paragraph{The Spatial Awareness Bottleneck.}
A sharp performance drop is observed when moving from perception to spatial reasoning. The average scores for Spatial Awareness are consistently lower than those for Perception (e.g., GPT-4o drops from 82.78 to 43.86), highlighting this dimension as a critical bottleneck. Tasks requiring precise geometric inference, such as \textit{2D Positioning} and \textit{Yaw Estimation}, prove exceptionally difficult, with many models scoring below 20.0. Notably, Gemini-2.0-Flash emerges as a significant outlier, achieving the highest average spatial score of 52.55 and outperforming GPT-4o by a large margin (8.69 points). This suggests that newer, efficient architectures may have retained superior geometric grounding capabilities.

\vspace{-0.5mm}

\paragraph{Commonsense and Functional Reasoning.}
In the Common Sense dimension, which tests the ability to map visual cues to user needs (e.g., identifying points of interest), we observe distinct behavior patterns. While models generally perform well in \textit{Single-View} identification, performance degrades in \textit{Cross-View} settings where spatial consistency is required. Once again, Gemini-2.0-Flash demonstrates dominance, particularly in Single-View reasoning (96.31), indicating a highly effective intent-to-location inference capability. However, the general decline in Cross-View scores across all models reinforces the finding that maintaining semantic consistency across viewpoint changes remains a major challenge for embodied agents.

\section{Conclusion}
We introduce ERGeoBench, a comprehensive evaluation framework designed to assess MLLM-based embodied agents across three progressively interactive settings and four fine-grained capability dimensions. Through extensive experiments, we identified key challenges, including difficulties in low-level metric localization and spatial reasoning, and the struggle to maintain semantic consistency across actively selected views. By highlighting these areas for improvement, we hope ERGeoBench will inspire and guide future research toward building more human-like and location-aware embodied agents.

\section*{Limitations}
A key limitation is that our evaluation is conducted in a panorama-based simulated environment rather than through physical deployment in live 3D environments. This reflects a trade-off between reproducibility, global coverage, and logistical feasibility. In particular, our current embodiment focuses on rotation, pitch, and zoom at a fixed panorama node; it does not yet evaluate translational movement between street nodes, obstacle-aware navigation, or dynamic scene interaction. While real-world testing is important for deployment, simulated benchmarks provide standardized and repeatable evaluation at global scale. Future work could incorporate continuous video streams, dynamic scenes, embodied movement over street graphs, and standardized real-world test suites to further bridge the gap between benchmark evaluation and practical navigation agents.

\section*{Acknowledgement}
This work is supported by  the National Key Research and Development Program of China under Grant 2024YFC3308500, Beijing Municipal Natural Science Foundation under Grant L251042, National Natural Science Foundation of China under Grant 62406036, China Postdoctoral Science Foundation under Grant 2025M781457,  and also sponsored by the State Key Laboratory of Networking and Switching Technology under Grant NST20250110.

\section*{Impact Statement}
This work aims to advance the evaluation of embodied geo-localization in multimodal large language models. The benchmark is intended to support safer and more reliable research on location-aware agents, but geo-localization technologies also raise privacy and misuse concerns, including unwanted inference of sensitive locations from visual content. ERGeoBench uses publicly available street-view-style panoramas and evaluates model capability rather than enabling identification of private individuals. We encourage future work to pair progress in geo-localization with privacy-preserving evaluation, misuse analysis, and safeguards against inferring sensitive personal locations.
\appendix

\bibliography{example_paper}
\bibliographystyle{icml2026}

\newpage
\appendix
\onecolumn
\section{Appendix Overview}
\label{app:overview}

This appendix provides implementation details and additional analyses that complement the main paper. It has two goals. First, it documents the unified prompt and agent protocol used for single-view, panorama-view, and embodied-view geo-localization. Second, it reports supplementary diagnostic analyses, including prompt refinement, annotation reliability, model-bias checks, panorama-versus-embodied comparison, and out-of-distribution robustness.

\begin{table}[H]
\centering
\caption{Organization of the appendix. The appendix emphasizes reproducibility of the prompt protocol and additional analyses that strengthen the interpretation of the main results.}
\label{tab:appendix_organization}
\small
\begin{tabular}{p{0.18\textwidth}p{0.74\textwidth}}
\toprule
\textbf{Section} & \textbf{Content} \\
\midrule
\S\ref{app:prompt_design} & Prompt design, mode-specific prompt variants, structured JSON schema, action protocol, and prompt-refinement ablation. \\
\S\ref{app:quality_control} & Data construction and annotation quality control, including manual filtering, model-assisted checking, and human verification. \\
\S\ref{app:cognitive_grounding} & Cognitive and neurological motivation for active evidence sampling, hypothesis updating, and spatial-memory consistency. \\
\S\ref{app:supp_experiments} & Supplementary experiments: GPT-5/o1 bias check, panorama-vs-embodied comparison, embodied exploration statistics, OOD wilderness evaluation. \\
\S\ref{app:repro_details} & Additional reproducibility details for rendering, inference, trajectory logging, and output parsing. \\
\bottomrule
\end{tabular}
\end{table}

\section{Prompt Design and Agent Protocol}
\label{app:prompt_design}

\subsection{Design Rationale}
\label{app:prompt_rationale}
The prompt is designed to turn geo-localization from one-shot visual recognition into a structured, evidence-seeking reasoning process.  Instead of directly asking the model to predict a location, the final prompt decomposes each step into four operations: structured observation, evidence evaluation, hypothesis update, and next-action planning.  This design is motivated by four recurring failure modes observed with a minimal prompt: single-cue overcommitment, premature guessing, weak uncertainty handling, and poor continuity across sequential views.

\begin{table}[H]
\centering
\caption{Prompt components and their intended function.}
\label{tab:prompt_components}
\small
\begin{tabular}{p{0.22\textwidth}p{0.68\textwidth}}
\toprule
\textbf{Component} & \textbf{Purpose} \\
\midrule
Multi-evidence integration & Prevents the model from deciding from a single clue, such as one billboard, one architectural cue, or one language fragment. The prompt requires multiple independent cues before high confidence. \\
No hallucination & Instructs the model not to invent text, flags, plate numbers, or landmarks. Unreadable evidence must be explicitly marked as unavailable or uncertain. \\
Progressive refinement & Encourages a coarse-to-fine process in which the model starts with a regional hypothesis and revises it when new evidence contradicts the previous belief. \\
Specific visual description & Replaces generic descriptions with concrete attributes such as roof type, lane marking, pole style, curb design, signage script, or vegetation density. \\
Consistency checking & Requires the model to compare climate, architecture, road infrastructure, traffic rules, and signage language against the predicted country or city. \\
\bottomrule
\end{tabular}
\end{table}

\subsection{Five-Category Observation Framework}
\label{app:five_category_prompt}

For consistency with the benchmark definition in the main paper, we organize visual evidence using the same 5-Slot Framework: Architecture, Signage, Vegetation, Terrain, and Infrastructure. This design keeps the prompt-level observation slots consistent with the benchmark taxonomy and the evaluated categories.

\begin{table}[H]
\centering
\caption{Five-slot observation framework used in the prompt and benchmark diagnostics.}
\label{tab:five_category_prompt}
\small
\begin{tabular}{@{}p{0.20\columnwidth}p{0.76\columnwidth}@{}}
\toprule
\textbf{Category} & \textbf{Diagnostic cues} \\
\midrule
Architecture & Building type, material, height, facade pattern, roof shape, age, density, and urban morphology. \\
Signage & Readable text, script/language, road signs, shop names, phone codes, domain suffixes, and brand hints. \\
Vegetation & Tree type, density, grass, aridity, seasonality, and climate-zone cues. \\
Terrain & Road surface, slope, visible landform, surrounding terrain type, openness, aridity, and other physical scene-layout cues. \\
Infrastructure & Driving side, lane markings, curb style, poles, insulators, sidewalks, guardrails, traffic lights, and plate color or shape. \\
\bottomrule
\end{tabular}
\end{table}

\subsection{Unified Mode-Specific Prompting}
\label{app:mode_specific_prompt}

The same output schema is used for all three visual settings.  The only prompt-level difference is a mode line that specifies whether the model can physically change the rendered view or should treat the action as an attention intention.  This keeps the reasoning protocol comparable across single-view, panorama-view, and embodied-view settings.

\begin{table}[H]
\centering
\caption{Mode-specific interpretation of the same prompt schema.}
\label{tab:mode_specific_prompt}
\small
\begin{tabular}{p{0.16\textwidth}p{0.32\textwidth}p{0.40\textwidth}}
\toprule
\textbf{Mode} & \textbf{Visual access} & \textbf{Interpretation of \texttt{next\_action}} \\
\midrule
Single-view & One fixed perspective view. & The model cannot change the viewpoint. The action is treated as an attention intention describing what it would inspect if it could move. \\
Panorama-view & One full equirectangular panorama. & The model receives the panorama at once. The action is treated as a region-of-interest attention plan over yaw, pitch, and zoom. \\
Embodied-view & Sequential egocentric views rendered from the panorama. & The model can control yaw, pitch, and zoom. The action is executed to obtain the next observation. \\
\bottomrule
\end{tabular}
\end{table}

\subsection{Structured Output Schema}
\label{app:json_schema}

\begin{figure*}[!b] \centering \includegraphics[width=0.95\textwidth]{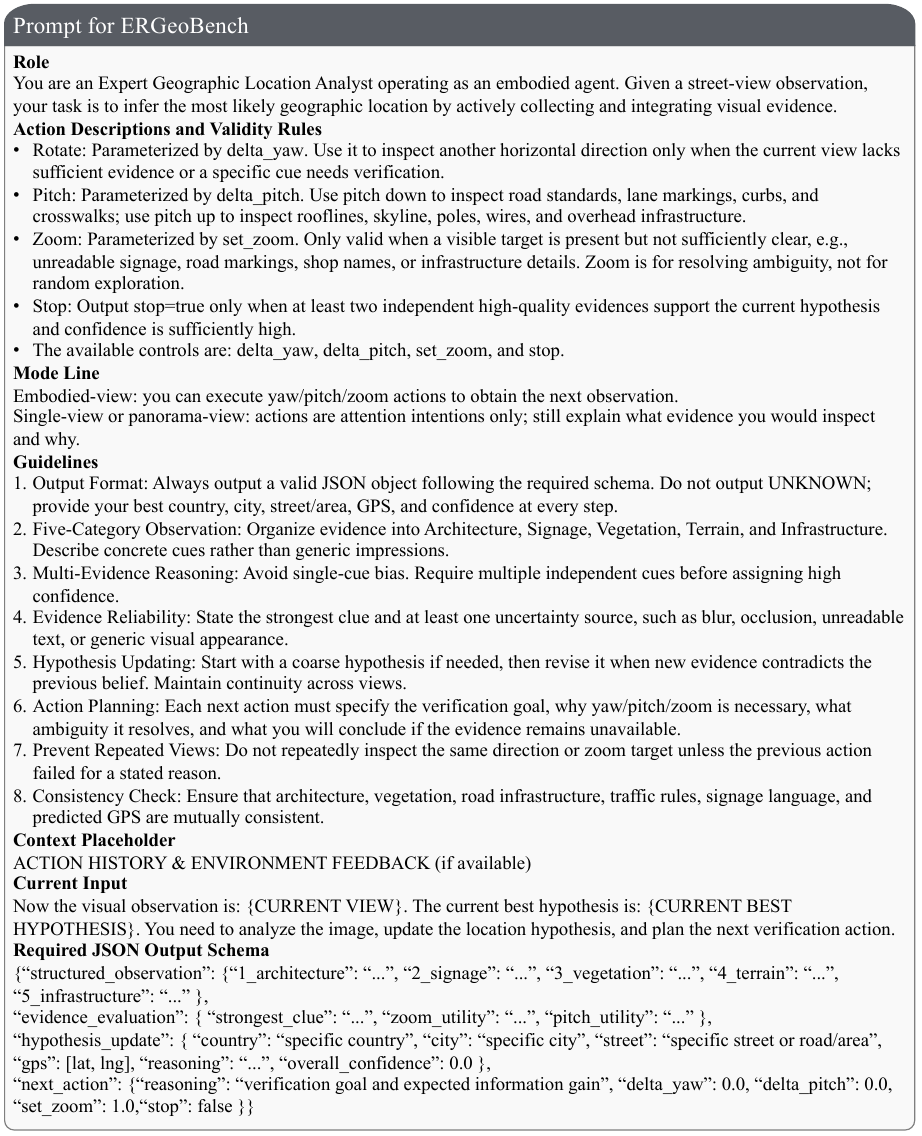} \caption{ Full prompt template used in ERGeoBench for embodied geo-localization. The prompt explicitly separates structured observation, evidence evaluation, hypothesis updating, and next-action planning to support active evidence acquisition and iterative reasoning. } \label{fig:prompt_template} \end{figure*}

The model is required to produce a JSON object with four fields: \texttt{structured\_observation}, \texttt{evidence\_evaluation}, \texttt{hypothesis\_update}, and \texttt{next\_action}, as shown in Fig.~\ref{fig:prompt_template}. The structured schema supports automatic parsing while also requiring the model to expose evidence, uncertainty, and the intended verification action.

\subsection{Verification-Oriented Action Protocol}
\label{app:action_protocol}

The prompt constrains actions to be verification-oriented rather than arbitrary exploration.  Yaw and pitch are used to search for additional evidence, while zoom is used only when a visible target is not sufficiently clear.  Typical action targets include readable street signs, shop names, language scripts, lane markings, curb standards, overhead wires, streetlight structures, and roofline or skyline features.  The prompt also specifies that pitch-down actions should be preferred for road-standard cues such as lane markings or crosswalks, while pitch-up actions should be used when overhead infrastructure or roof-mounted features may disambiguate the region.

At step $t$, the model receives the current observation, a short history of recent views, and the current best hypothesis.  It then predicts a new hypothesis and either stops or emits $\Delta\psi$, $\Delta\phi$, and a zoom level for the next view.  The run terminates when the model chooses to stop or when the confidence is sufficiently high.  Conceptually, this implements a closed loop:
\begin{equation}
\label{eq:closed_loop_appendix}
I_t, H_t, \hat{y}_{t-1} \rightarrow \left(o_t, e_t, \hat{y}_t, a_t\right),
\end{equation}
where $I_t$ is the current observation, $H_t$ is the recent view history, $\hat{y}_{t-1}$ is the previous hypothesis, $o_t$ is the structured observation, $e_t$ is the evidence evaluation, $\hat{y}_t$ is the updated location hypothesis, and $a_t$ is the next action.

\subsection{Prompt Refinement Ablation}
\label{app:prompt_ablation}

We compare the final structured prompt against a minimal base prompt.  The base prompt directly asks the model to geolocate the image and output country, city, street/area, coordinates, and confidence.  The final prompt adds five-category scene decomposition, explicit evidence evaluation, step-wise hypothesis updating, and verification-oriented actions.

\begin{table}[H]
\centering
\caption{Prompt refinement ablation. The structured prompt substantially improves semantic accuracy and reduces distance error across all three settings.}
\label{tab:prompt_refinement_ablation}
\small
\begin{tabular}{llcccc}
\toprule
\textbf{Setting} & \textbf{Prompt} & \textbf{City Acc.} & \textbf{Country Acc.} & \textbf{Acc.@25km} & \textbf{Avg. Error (km)} \\
\midrule
Single-view & Base & 16.04 & 32.67 & 38.70 & 2044.62 \\
Single-view & Ours & \textbf{46.21} & \textbf{84.16} & \textbf{63.91} & \textbf{662.18} \\
\midrule
Panorama-view & Base & 11.06 & 20.57 & 29.63 & 2049.31 \\
Panorama-view & Ours & \textbf{54.03} & \textbf{92.71} & \textbf{75.50} & \textbf{231.51} \\
\midrule
Embodied-view & Base & 24.62 & 72.67 & 52.44 & 2492.40 \\
Embodied-view & Ours & \textbf{55.08} & \textbf{92.33} & \textbf{76.10} & \textbf{224.05} \\
\bottomrule
\end{tabular}
\end{table}

\section{Data Construction and Annotation Quality Control}
\label{app:quality_control}

\subsection{Multi-Stage Quality-Control Pipeline}
\label{app:quality_control_pipeline}

ERGeoBench uses model assistance for scalable construction, but the final annotations are not produced by a single model in a one-shot manner.  The construction pipeline combines manual filtering, model-assisted generation, independent consistency checking, and human verification.  Incorrect, ambiguous, or visually unreliable samples are corrected, discarded, or regenerated.

\begin{table}[H]
\centering
\caption{Quality-control stages used in ERGeoBench construction.}
\label{tab:quality_control_pipeline}
\small
\begin{tabular}{p{0.18\textwidth}p{0.72\textwidth}}
\toprule
\textbf{Stage} & \textbf{Role} \\
\midrule
Manual panorama filtering & Remove unsuitable panoramas, including indoor scenes, severely blurred images, low-resolution views, and panoramas heavily occluded by the street-view vehicle. \\
Target view declaration & Ground each question or annotation in a specified field of view so that the intended evidence is visible and spatially localized. \\
Model-assisted generation & Generate candidate questions, answers, or evidence descriptions with an MLLM to scale annotation. \\
Independent consistency check & Use an independent VLM checker to identify mismatches between the candidate annotation and the visual evidence. \\
Human verification & Manually inspect final samples and remove, correct, or regenerate inconsistent or ambiguous cases. \\
\bottomrule
\end{tabular}
\end{table}

\subsection{Ground Truth for Functional and Spatial Tasks}
\label{app:ground_truth_definition}

For common-sense reasoning tasks, the ground truth is represented as a discrete point-of-interest or functional label, and evaluation uses fuzzy text matching to allow minor naming variations.  For spatial-awareness tasks, ground truth is represented using constrained natural-language answers, including coarse depth bins and discrete comparative outcomes such as left/right, front/back, and cross-view object relations.  A judge model then applies explicit matching rules.  We do not define reachability using pixel-level thresholds, 3D reconstruction, or path-planning distance, because ERGeoBench is designed as a reproducible panorama-based diagnostic benchmark rather than a full physical navigation simulator.

\section{Cognitive and Neurological Motivation}
\label{app:cognitive_grounding}

Embodied geo-localization is naturally an active perception problem.  Humans rarely localize from a single static glance; instead, they sample the scene, attend to diagnostic cues, maintain a spatial representation, and update hypotheses as new evidence becomes available.  ERGeoBench operationalizes this process through a closed-loop protocol that asks the model to observe, evaluate evidence, update its hypothesis, and plan a targeted next action.  This design is consistent with research showing that eye movements in natural behavior are shaped by task goals \cite{hayhoe2005eye}, that agents actively sample information under curiosity or expected-utility considerations \cite{gottlieb2018towards}, and that navigation depends on cognitive maps, place/grid representations, head-direction signals, and scene-memory networks \cite{epstein2017cognitive,moser2008place,taube2007head,steel2021network}.

\begin{table}[H]
\centering
\caption{Mapping between cognitive motivation and benchmark implementation.}
\label{tab:cognitive_mapping}
\small
\begin{tabular}{p{0.24\textwidth}p{0.26\textwidth}p{0.40\textwidth}}
\toprule
\textbf{Principle} & \textbf{Implementation in ERGeoBench} & \textbf{Expected model capability} \\
\midrule
Active information sampling & The agent selects yaw, pitch, and zoom actions to verify specific clues. & Choose informative views rather than passively consuming a fixed image. \\
Spatial representation and memory updating & The prompt includes recent view history and a current best hypothesis. & Maintain cross-view continuity and revise beliefs as new evidence arrives. \\
Goal-conditioned affordance reasoning & The next action must state a concrete verification goal and expected ambiguity reduction. & Decide what evidence is worth inspecting for localization. \\
Multi-cue integration & The model must organize evidence into structured categories and identify uncertainty. & Avoid single-cue bias and hallucinated fine-grained predictions. \\
\bottomrule
\end{tabular}
\end{table}

\section{Supplementary Experiments and Analyses}
\label{app:supp_experiments}

\subsection{Additional OpenAI Model Evaluation for Bias Checking}
\label{app:gpt_bias_check}

To test whether the benchmark favors GPT-4o because GPT-4o is involved in candidate annotation, we additionally evaluate GPT-5 and o1 under the same protocol.  The results do not support a GPT-4o-specific advantage: GPT-5 outperforms GPT-4o in all three geo-localization settings, while o1 also exceeds GPT-4o in single-view geo-localization.

\begin{table}[H]
\centering
\caption{Geo-localization performance of GPT-4o, GPT-5, and o1.}
\label{tab:gpt5_o1_geoloc}
\small
\begin{tabular}{llc}
\toprule
\textbf{Model} & \textbf{Setting} & \textbf{GLS} $\uparrow$ \\
\midrule
GPT-4o & Single-view & 61.98 \\
GPT-5 & Single-view & 64.13 \\
o1 & Single-view & \textbf{65.60} \\
\midrule
GPT-4o & Panorama-view & 67.06 \\
GPT-5 & Panorama-view & \textbf{67.95} \\
o1 & Panorama-view & 66.71 \\
\midrule
GPT-4o & Embodied-view & 67.55 \\
GPT-5 & Embodied-view & \textbf{69.54} \\
o1 & Embodied-view & 67.43 \\
\bottomrule
\end{tabular}
\end{table}

\begin{table}[H]
\centering
\caption{Capability-level comparison of GPT-4o, GPT-5, and o1. GPT-5 improves over GPT-4o on all three diagnostic dimensions.}
\label{tab:gpt5_o1_capability}
\small
\begin{tabular}{lccc}
\toprule
\textbf{Dimension} & \textbf{GPT-4o} & \textbf{GPT-5} & \textbf{o1} \\
\midrule
Foundational Perception & 82.78 & \textbf{89.81} & 68.39 \\
Spatial Awareness & 43.86 & \textbf{64.90} & 55.63 \\
Common Sense & 42.38 & \textbf{53.96} & 48.37 \\
\bottomrule
\end{tabular}
\end{table}

\subsection{Value of Embodied Evaluation Beyond Panorama Evaluation}
\label{app:embodied_vs_panorama}

Panorama-view evaluation provides near-complete visual access at once, whereas embodied-view evaluation measures active viewpoint selection, temporal evidence accumulation, and spatial consistency under sequential egocentric observations.  The embodied setting is therefore not intended to replace panorama evaluation; it diagnoses additional agentic abilities.  Quantitatively, several strong models match or exceed their panorama GLS in the embodied setting, and fine-grained Acc.@1km improves consistently for the models shown below.

\begin{table}[H]
\centering
\caption{Quantitative comparison between panorama-view and embodied-view evaluation.}
\label{tab:panorama_embodied_comparison}
\small
\begin{tabular}{lcccc}
\toprule
\textbf{Model} & \textbf{Panorama GLS} & \textbf{Embodied GLS} & \textbf{$\Delta$ GLS} & \textbf{Acc.@1km: Pan $\rightarrow$ Emb} \\
\midrule
GPT-4o & 67.06 & 67.55 & +0.49 & 7.58 $\rightarrow$ 9.73 \\
Gemini-2.0-Flash & 66.18 & 67.13 & +0.95 & 7.15 $\rightarrow$ 9.09 \\
Gemini-2.5-Pro & 68.28 & 69.02 & +0.74 & 15.97 $\rightarrow$ 16.75 \\
Gemini-3-Flash & 68.19 & 68.81 & +0.62 & 20.87 $\rightarrow$ 22.96 \\
\bottomrule
\end{tabular}
\end{table}

\subsection{Out-of-Distribution Wilderness Stress Test}
\label{app:ood_wilderness}

ERGeoBench mainly targets representative urban geo-localization.  To test sparse-cue robustness, we additionally evaluate a small OOD set containing 20 extreme wilderness scenes.  All tested models show large GLS drops, indicating that wilderness scenes remain challenging even for strong proprietary MLLMs.

\begin{table}[H]
\centering
\caption{GLS comparison between the standard benchmark and the OOD wilderness stress test.}
\label{tab:ood_wilderness}
\small
\begin{tabular}{lccc}
\toprule
\textbf{Model} & \textbf{Standard Benchmark} & \textbf{OOD Set} & \textbf{Drop} \\
\midrule
GPT-4o & 67.55 & 42.61 & 24.94 \\
Gemini-2.5-Pro & 69.02 & 47.81 & 21.21 \\
Qwen-VL-Max & 42.54 & 22.22 & 20.32 \\
\bottomrule
\end{tabular}
\end{table}

\section{Additional Reproducibility Details}
\label{app:repro_details}

\subsection{Rendering and Inference Configuration}
\label{app:rendering_config}

All three settings are implemented with a unified script and shared structured-output schema.  The single-view setting renders a fixed perspective view, the panorama-view setting sends a scaled panorama or a zoomed panorama crop, and the embodied-view setting renders sequential perspective observations from the underlying panorama.  Unless otherwise specified, inference uses temperature $0.1$ and a maximum generation length of $4096$ tokens.  Panorama images are optionally resized so that the long side does not exceed $1800$ pixels to reduce token cost, and JPEG quality is set to $92$ for model input.

\subsection{Trajectory Logging}
\label{app:trajectory_logging}

For embodied evaluation, each trajectory record stores the step index, yaw, pitch, zoom, structured observation, complete JSON output, and optional saved image name.  The recent trajectory history is summarized and passed back into the next prompt, enabling the model to reason over visual continuity rather than treating each observation independently.  The output trajectory is serialized as JSON for downstream evaluation and reproducibility.

\subsection{Parsing and Invalid Outputs}
\label{app:parsing_invalid}

Model outputs are parsed by extracting the first valid JSON object from the response.  Predictions with missing location fields, placeholder coordinates, or forbidden unknown-like answers are treated as invalid under the single-pass benchmark protocol.  This design intentionally evaluates not only visual geo-localization ability, but also whether a model can reliably follow a structured response protocol under multimodal reasoning constraints.

\end{document}